\documentclass{article}

\usepackage[utf8]{inputenc}
\usepackage[T1]{fontenc}

\usepackage{amsmath}
\usepackage{amssymb}
\usepackage{amsthm}
\usepackage{mathtools}

\usepackage{graphicx}
\usepackage{float}

\usepackage{hyperref}
\usepackage{url}
\usepackage{natbib}

\newtheorem{theorem}{Theorem}

\theoremstyle{definition}
\newtheorem{definition}[theorem]{Definition}
\theoremstyle{remark}

\usepackage[margin=1in]{geometry}

\title{WavePhaseNet: A DFT-Based Method for Constructing Semantic Conceptual Hierarchy Structures (SCHS)}

\author{
  Kiyotaka Kasubuchi \\
  \texttt{[affiliation]} \\
  \and
  Kazuo Fukiya \\
  \texttt{[affiliation]}
}

\date{April 12, 2026}

\begin{document}

\maketitle

\begin{abstract}
In this paper, we reformulate the Transformer/Attention mechanisms in Large Language Models (LLMs) from the perspectives of measure theory and frequency analysis, and theoretically clarify the inevitability of hallucination as a structural limitation. In particular, we focus on the fact that the embedding space functions as a conditional expectation over a $\sigma$-algebra, and demonstrate that the failure of this framework to be isomorphic to the semantic truth set is the fundamental cause of the breakdown of logical consistency.

Based on this observation, we propose \textbf{WavePhaseNet}, a method for explicitly constructing a \textbf{Semantic Conceptual Hierarchy Structure (SCHS)} within the embedding space using the \textbf{Discrete Fourier Transform (DFT)}. By applying a DFT along the sequence dimension, meaning is decomposed into frequency bands: low frequencies are progressively separated as global meaning and intent, and high frequencies as local syntax and expression. This makes it possible to carry out semantic manipulation rigorously in a diagonalized space.

As a concrete example, we show that the 24{,}576-dimensional high-dimensional embedding space, typified by GPT-4, exhibits a $1/f$ spectral structure based on the self-similarity of language and Zipf's law. Through cumulative energy analysis, we theoretically derive that the lower bound of ``complete representation'' is approximately 3{,}000 dimensions. This demonstrates that dimensionality reduction from 24{,}576 to 3{,}000 dimensions is possible while preserving meaning and intent, thereby enabling rigorous reasoning while suppressing hallucination.

Finally, since the reduced low-dimensional embedding space is constructed by combining a cohomological regularization principle over overlapping local windows, a graph structure can be introduced over the window covering, and a cochain complex can be defined, so that inconsistencies among local inferences can be quantified as a loss based on coboundaries. By applying harmonic projection based on Hodge theory to extract a maximally consistent global representation, this positions cohomology not as a purely topological invariant, but as a computable regularization principle for controlling semantic consistency. The knowledge-reasoning process thereby maintains validity and rigor through the gluing of knowledge fragments.
\end{abstract}

\section{Introduction}

Transformer-based LLMs realize natural language generation as \textbf{autoregressive (AR) models} by sequentially approximating the conditional probability distribution of a token sequence. While this framework demonstrates extremely high performance in grammatical consistency and expressive diversity, it inevitably produces erroneous generations---so-called \emph{hallucinations} (false outputs)---in scenarios that demand a high degree of logical reasoning or factual consistency.

Reasoning in LLMs relies primarily on locally computed intermediate representations, and global semantic consistency is guaranteed only implicitly. Although the self-attention mechanism can handle long-range dependencies, it lacks any mechanism to directly suppress the semantic inconsistencies that arise between overlapping local contexts.

The purpose of this paper is to position hallucination not as a mere consequence of insufficient training or data defects, but as an inevitable phenomenon arising from the mathematical structure of the model itself, and to provide, as its solution, a reconstruction principle for the embedding space.

Meanwhile, spectral analysis of token-embedding sequences has suggested that global semantic structure and local contextual/syntactic structure can be separated (see the prior work FNet; \citealp{fnet}).

In this paper, we integrate these two perspectives and propose a new inference framework that combines spectrally extracted global intent with a cohomology-inspired local-consistency regularization.

Concretely, we introduce cochains/cohomology over a covering of local window groups together with a graph-based Hodge decomposition (Laplacian projection), and construct a method for \emph{semantic gluing} that incorporates the inconsistency (coboundary) among local inferences (sections) into the loss. In practice, this can be implemented by introducing a Spectral Module and a Cohomology Regularizer into each layer and each head of a Transformer.

\paragraph{First concept.}
By generating, via DFT, the token-embedding matrix as a \emph{semantic-hierarchical token-embedding matrix} in which global themes/intents and local syntactic information are mixed, the representation is mapped and analyzed from syntactic similarity toward a semantic neighborhood system.

\paragraph{Second concept.}
By \emph{reducing} the resulting semantic-hierarchical token-embedding matrix into semantic latent variables, we achieve a compact, low-cost semantic rigor while preserving semantic concepts.

\paragraph{Third concept.}
We show that a semantic knowledge-reasoning process is obtained by \textbf{harmonic gluing}: among the set of local representations that satisfy the task loss, one selects the representation that minimizes coboundary energy and is projected onto the harmonic subspace of the graph Laplacian induced by the window covering.

\section{Autoregressive Models and the Embedding Space as a $\sigma$-algebra}

\subsection{Measure-Theoretic Formulation of the Autoregressive Model}

An autoregressive model is the operation of decomposing the joint distribution $P(X_1,\ldots,X_N)$ as
\begin{equation}
P(X_1,\ldots,X_N) = \prod_{t=1}^{N} P(X_t \mid X_{<t}).
\end{equation}
Here, the information set generated by the past token sequence $X_{<t}$ (a Borel set) forms the $\sigma$-algebra
\begin{equation}
\mathcal{F}_t = \sigma(x_1,\ldots,x_t),
\end{equation}
and this chained regression model takes the form
\begin{equation}
P(x_{t+1} \mid \mathcal{F}_t).
\end{equation}

Thus, an embedding vector $V$ on the measurable space $(\Omega, \mathcal{F}_t, P)$ can be interpreted as a \textbf{random variable over the $\sigma$-algebra}.

Consequently, the LLM as a whole is a \emph{sequential approximator of conditional expectations over $\sigma$-algebras}, and its output is not a pointwise truth value but an average optimality with respect to a probability distribution (a Lebesgue measure: the expectation $E$).

\subsection{Norm Structure of $L^p$ Spaces}

The embedding space is treated, by way of an inner-product space, as a normed space representing the divergence between the truth manifold and the support of the learning distribution.

\begin{itemize}
\item Qualitatively, each layer approximates
\begin{equation}
E[\phi(X_{t+1}) \mid \mathcal{F}_t].
\end{equation}
\item Since the expected value of the loss function is
\begin{equation}
E[l(X_{t+1}, \hat{X}_{t+1})],
\end{equation}
the model as a whole resides in
\begin{equation}
L^p(\Omega, \mathcal{F}, P).
\end{equation}
\item This \textbf{norm of the $L^p$ space} is equivalent to the Lebesgue integral, i.e.\ the expectation of the probability distribution,
\begin{equation}
\|f\|_p = \left(\int_{\Omega} |f(\omega)|^p \, dP(\omega)\right)^{1/p},
\end{equation}
\end{itemize}
and this expression shows that what is optimized is not a pointwise truth value but the \emph{plausible average optimality} of a distribution.

\begin{figure}[H]
\centering
\includegraphics[width=0.6\textwidth]{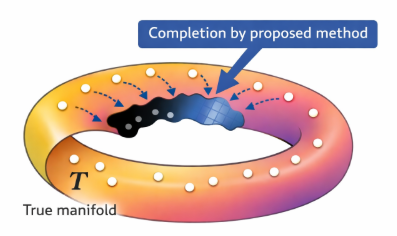}
\caption{Conceptual diagram of the manifold: completing the defects (holes) on the truth manifold $T$.}
\end{figure}

\subsection{Norms and the Minkowski Inequality}

In an $L^p$ space, the Minkowski inequality
\begin{equation}
\|f + g\|_p \leq \|f\|_p + \|g\|_p
\end{equation}
holds. Its meaning and interpretation are:
\begin{itemize}
\item Different semantic components, hypotheses, and contexts can be linearly composed in the embedding space, and
\item moreover, the norm does not ``break down'' (there is no gradient explosion).
\end{itemize}

This is a property that permits \textbf{semantic superposition}, and it is a source of the flexibility of generative AI.

On the other hand, the combination of this linearity with incompleteness inherently contains a structure capable of generating non-existent propositions (erroneous outputs).

\section{Mathematical Definition of Hallucination}

\subsection{Definition}

We define hallucination as follows.

\begin{definition}[Hallucination]
An output that is optimal (high-probability) over the $\sigma$-algebra, but does not belong to the truth set of the real world.
\end{definition}

Mathematically, with
\begin{itemize}
\item the truth set $T \subset V^{*}$,
\item the set the model optimizes over $\displaystyle \arg\min_{f} \mathbb{E}\left[l(f(X),Y)\right]$,
\end{itemize}
the truth value of general knowledge satisfies
\begin{equation}
T \neq \arg\min L^{p}.
\end{equation}

By the Minkowski inequality: although two distinct hypotheses $f_1, f_2$ can be linearly composed as $f = \alpha f_1 + (1-\alpha) f_2$ in the probability-distribution space $L^p$, incompleteness allows non-existent propositions to arise.

\subsection{What Hallucination in LLMs Is}

Hallucination in LLMs is an \textbf{inevitable phenomenon arising from the structural mismatch that ``the truth set is nonlinear, whereas the $L^p$ space is convex.''}

Moreover, the statistical averaging (the ``law of large numbers'') used by the Transformer only exhibits a ``distribution that is plausible on average''; AR plus gradient optimization is likewise insufficient, and therefore \textbf{\v{C}ech-type truth consistency is necessarily lost}.

\begin{itemize}
\item The truth set is in general nonconvex and is not closed under linear combination, whereas the embedding space has a convex structure.
\item It is precisely this mismatch between the ``nonlinear truth set'' and the ``linear, convex representation space'' that is the essential cause of hallucination.
\item Hence, while the model is well suited to high-degree-of-freedom, diverse tasks such as sentence generation, it is ill-suited to reasoning processes that require rigorous logical development. This is a fatal defect for AI agents and edge computing.
\end{itemize}

\section{The Necessity of the Semantic Conceptual Hierarchy Structure (SCHS)}

\subsection{Overview}

\begin{itemize}
\item Token-level AR (grammatical consistency)
\begin{itemize}
\item is turned into frequency-level AR (semantic-hierarchical consistency).
\end{itemize}
\item DFT autoregression:
\begin{equation}
P(\hat{X}) = p\!\left(\hat{X}_{\text{low}}\right)\prod_{k > \text{low}} p\!\left(\hat{X}_{k} \mid \hat{X}_{<k}\right),
\end{equation}
i.e.\ SCHS formation by semantic-resolution decomposition.
\end{itemize}

In natural language processing (NLP), knowledge-structure theory, graph theory, predicate logic, and so on, diverse semantic-concept \textbf{inheritance} is expressed by a ``hierarchical structure'' over the semantic-concept space:
\begin{itemize}
\item Thesaurus: the hierarchical structure of super-/sub-ordinate concepts.
\item Ontology: the hierarchical structure of concept classes, individuals, attributes, and relations.
\item Semantic network: the network structure of inheritance (is-a) and part-of (has-a).
\end{itemize}

In a Transformer, super-/sub-ordinate concepts, concept classes, attributes, and relations can be sought by adding \emph{role tokens}, but this does not mean that a concept hierarchy is explicitly intrinsic to the space.

The output of the attention mechanism does not take into account the super-ordinate concepts or intent of the input prompt; it is merely token generation estimated from the neighborhood of a direction-vector point to which ``role tokens'' have been added.

Therefore, when asking a somewhat complex reasoning question, the current situation is that advanced linguistic skill and problem-structuring ability are demanded of the user. This is also the reason why \emph{prompt engineering} has become a necessity.

\subsection{Generation of the SCHS}

The semantic-concept space is composed of a continuous, smooth topological manifold; but the embedding vectors of the embedding space that models it are discrete vectors. By transforming these into continuous, distinct frequencies, a smooth hierarchical structure of semantic concepts is obtained.

However, this form is hard to handle, so by \emph{inverse-transforming} it back into a discrete matrix, we obtain a tensor (matrix) into which the concept hierarchy is explicitly embedded and which the attention mechanism can handle directly. The core of this proposal is that, before the inverse transform, the intrinsic dimensionality is reduced---e.g.\ from GPT-4's 24{,}576 dimensions to the 3{,}000-dimensional lower bound of complete representation---thereby suppressing hallucination and enabling a rigorous reasoning process even for complex, redundant inputs.

Input: the embedding matrix of a token sequence,
\begin{equation}
X \in \mathbb{R}^{d \times N} \quad (\text{embedding dimension } d,\ \text{sequence length } N).
\end{equation}
For a sequence of sentence length $N$,
\begin{equation}
x = (x_0, x_1, \ldots, x_{N-1}), \quad x_i \in \mathbb{R}^{d}.
\end{equation}
The discrete Fourier transform (DFT) along the token-sequence direction is defined by
\begin{equation}
\hat{x} = (\hat{x}_0, \hat{x}_1, \ldots, \hat{x}_{N-1}), \quad \hat{x}_i \in \mathbb{C}^{d}.
\end{equation}

\subsubsection{Sequence-Direction DFT (per each of the $d$ embedding dimensions)}

\begin{equation}
\hat{X}_{k} = \sum_{n=0}^{N-1} x_{n}\, e^{-2\pi i \frac{kn}{N}}
\qquad \left(x_{n} \in X;\ n = 0,\ldots,N-1,\ k = 0,\ldots,N-1\right),
\end{equation}
where $\hat{X}_{k} \in \mathbb{C}^{d}$ is the complex exponential vector at frequency $k$.

In matrix form,
\begin{equation}
\hat{X} = X F_{N}^{\top} \qquad (F \text{ is the } N \times N \text{ DFT matrix}),
\end{equation}
so that $\hat{X} \in \mathbb{C}^{d \times N}$, and each entry is transformed as
\begin{equation}
\hat{X}_{j,k} = \sum_{n=0}^{N-1} X_{j,n}\, e^{-2\pi i \frac{kn}{N}}.
\end{equation}

We illustrate the construction of the DFT matrix $F \in \mathbb{C}^{N \times N}$ with the example $N = 4$. As a sample computation of the $(4,4)$ entry of the square matrix $[(\text{column } k = 4:\ 0,1,2,3) \times (\text{row } N = 4 \text{ tokens})]$, at $(k,n) = (3,3)$:
\begin{equation}
\omega_{3}^{3} = \omega^{9} = e^{-2\pi i \frac{3\times 3}{4}} = e^{-\frac{18\pi i}{2}} = e^{-\frac{\pi i}{2}} = \cos\!\left(-\tfrac{\pi}{2}\right) + i\sin\!\left(-\tfrac{\pi}{2}\right) = -i.
\end{equation}
Hence the DFT matrix is
\begin{equation}
F_{4} = \begin{bmatrix}
1 & \cdots & 1 \\
\vdots & \ddots & \vdots \\
1 & \cdots & \omega^{9}
\end{bmatrix}
= \begin{bmatrix}
1 & \cdots & 1 \\
\vdots & \ddots & \vdots \\
1 & \cdots & -i
\end{bmatrix}.
\end{equation}
That is, as many frequencies as there are tokens are derived, yielding the four-fold cyclic analysis $(1, i, -1, -i)$. Transposing,
\begin{equation}
F_{4}^{\top} = \begin{bmatrix}
1 & \cdots & 1 \\
\vdots & \ddots & \vdots \\
1 & \cdots & -i
\end{bmatrix}^{\top}
= \begin{bmatrix}
1 & \cdots & 1 \\
\vdots & \ddots & \vdots \\
1 & \cdots & -i
\end{bmatrix}.
\end{equation}
Multiplying by this transposed DFT matrix decomposes the ``variation along the token sequence'' frequency by frequency, for each embedding dimension.

\subsubsection{Amplitude--Phase Decomposition (Complex Representation)}

For each component, the DFT layer (FFT) gives
\begin{equation}
\hat{X}_{k} = A_{k} \odot e^{i \Phi_{k}},
\end{equation}
with
\begin{align}
\text{phase:}\quad & \Phi_{k} = \angle \hat{X}_{k} \in \mathbb{R}^{d}, \\
\text{amplitude:}\quad & A_{k} = |\hat{X}_{k}| \in \mathbb{R}^{d}.
\end{align}
The inverse transform (reconstruction) is
\begin{equation}
x_{n} = \frac{1}{N}\sum_{k=0}^{N-1} \hat{X}_{k}\, e^{2\pi i \frac{kn}{N}}.
\end{equation}

\paragraph{Frequency-band partition.}
Let the low-frequency set be $L = \{k : |k| \leq K_{\text{low}}\}$; the low frequencies carry the long-period band (the topic/summary and semantic analysis of the whole sentence). The high-frequency set $H$ is the remainder (local syntax/context analysis; in practice defined by taking the symmetry of the DFT into account), and the high frequencies mainly carry the rapidly varying components (pragmatics, expression, and the fine details of generation; syntax/context analysis). This rests on the NLP assumption stated above. Below, we prove this assumption using a simple worked example.

\subsection{A Simple Numerical Example ($d = 4$, $N = 4$): What the DFT Does (Actual Numbers)}

Input (4 tokens, embedding dimension 4). Example sentence: ``Japan's / capital / is / Tokyo.''
\begin{equation}
X = \begin{bmatrix}
1.0 & 0.5 & -0.2 \\
0.0 & 1.0 & 0.3 \\
0.5 & -0.5 & 0.0 \\
1.5 & 0.0 & 0.1
\end{bmatrix} \tag{1}
\end{equation}

Computing the sequence-direction DFT (each row is the coefficient of frequency $k = 0,1,2,3$) gives, as complex numbers,
\begin{equation}
\hat{X} = \begin{bmatrix}
3.0 + 0.0i & 1.0 + 0.0i & 0.2 + 0.0i \\
0.5 + 1.5i & 1.0 - 1.0i & -0.2 - 0.2i \\
0.0 + 0.0i & -1.0 + 0.0i & -0.6 + 0.0i \\
0.5 - 1.5i & 1.0 + 1.0i & -0.2 + 0.2i
\end{bmatrix} \tag{2}
\end{equation}
substituting into
$\hat{X}_{k} = \sum_{n=0}^{N-1} x_{n}\, e^{-2\pi i \frac{kn}{N}}$.

A specific computation:
\begin{align}
\hat{X}_{0} &= \sum_{n=0}^{3} x_{n}\, e^{-2\pi i \frac{0\cdot n}{4}}
= \sum_{n=0}^{3} x_{n}\cos\!\left(\tfrac{\pi\, 0\, n}{2}\right) - i\sum_{n=0}^{3} x_{n}\sin\!\left(\tfrac{\pi\, 0\, n}{2}\right) \\
&= (x_0 + x_1 + x_2 + x_3) + i(0.0) \\
&= (1.0, 0.5, -0.2) + (0.0, 1.0, 0.3) + (0.5, -0.5, 0.0) + (1.5, 0.0, 0.1) + i(0.0) \\
&= (3.0, 1.0, 0.2) + i(0.0),
\end{align}
which is row 1 of (2): $[\,3.0+0.0i\ \ 1.0+0.0i\ \ 0.2+0.0i\,]$.
\begin{align}
\hat{X}_{1} &= \sum_{n=0}^{3} x_{n}\, e^{-2\pi i \frac{n}{4}}
= \sum_{n=0}^{3} x_{n}\cos\!\left(\tfrac{\pi n}{2}\right) - i\sum_{n=0}^{3} x_{n}\sin\!\left(\tfrac{\pi n}{2}\right)
= (x_0 - x_2) - i(x_1 - x_3) \\
&= (1.0, 0.5, -0.2) - (0.5, -0.5, 0.0) - i(0.0, 1.0, 0.3) + i(1.5, 0.0, 0.1) \\
&= (0.5, 1.0, -0.2) + i(1.5, -1.0, -0.2),
\end{align}
which is row 2 of (2): $[\,0.5+1.5i\ \ 1.0-1.0i\ \ -0.2-0.2i\,]$. The components $\hat{X}_{2}, \hat{X}_{3}$ are obtained in the same way.

The amplitude $A$ (modulus) and phase $\Phi$ (argument):
\begin{equation}
A_{k,\cdot} = \begin{bmatrix}
3.0 & 1.0 & 0.2 \\
1.5811 & 1.4142 & 0.2828 \\
0.0 & 1.0 & 0.6 \\
1.5811 & 1.4142 & 0.2828
\end{bmatrix} \tag{3}
\end{equation}
\begin{equation}
\Phi_{k,\cdot} = \begin{bmatrix}
0 & 0 & 0 \\
1.2490 & -0.7854 & -2.3562 \\
0 & \pi & \pi \\
-1.2490 & 0.7854 & 2.3562
\end{bmatrix} \tag{4}
\end{equation}

In this example, note that $k=0$ (the DC term) is $[3, 1, 0.2]$ and represents the global feature (summary, intent) of the whole sentence; $k = 1, 3$ are the conjugate pair.

A specific computation---amplitude $A$: the $(2,1)$ entry (row 2, column 1) of $A_{k,\cdot}$ derived from (2) is
\begin{equation}
\sqrt{0.5^{2} + 1.5^{2}} = \sqrt{0.25 + 2.25} = \sqrt{2.5} = 1.58113883\cdots \cong 1.5811.
\end{equation}
Phase $\Phi$: the $(2,1)$ entry of $\Phi_{k,\cdot}$ derived from (2) is
\begin{equation}
\operatorname{atan2}(1.5, 0.5) = \arctan\!\left(\tfrac{1.5}{0.5}\right) = \arctan(3) = 1.24904577\cdots \cong 1.2490.
\end{equation}
The other entries are obtained similarly.

\subsubsection{Definition of the Inverse DFT (Sequence Direction)}

For positions $n = 0,1,2,3$,
\begin{equation}
x_{n} = \frac{1}{N}\sum_{k=0}^{N-1} \hat{X}_{k}\, e^{2\pi i \frac{kn}{N}} \qquad (N = 4),
\end{equation}
where
\begin{itemize}
\item $x_{n} \in \mathbb{R}^{3}$ is the embedding at position $n$,
\item $\hat{X}_{k} \in \mathbb{C}^{3}$ is the coefficient at frequency $k$.
\end{itemize}
The exponential factor is
\begin{equation}
e^{2\pi i \frac{kn}{4}} = \begin{cases}
1 & (kn = 0), \\
i & (kn \equiv 1 \bmod 4), \\
-1 & (kn \equiv 2 \bmod 4), \\
-i & (kn \equiv 3 \bmod 4).
\end{cases}
\end{equation}

\subsubsection{Given Frequency Coefficients (Restated)}

\begin{align}
\hat{X}_{0} &= (3.0,\ 1.0,\ 0.2), \\
\hat{X}_{1} &= (0.5 + 1.5i,\ 1.0 - 1.0i,\ -0.2 - 0.2i), \\
\hat{X}_{2} &= (0.0,\ -1.0,\ -0.6), \\
\hat{X}_{3} &= (0.5 - 1.5i,\ 1.0 + 1.0i,\ -0.2 + 0.2i).
\end{align}
(Since $\hat{X}_{3} = \overline{\hat{X}_{1}}$, real reconstruction is guaranteed.)

\subsubsection{Inverse Transform at Each Position}

\paragraph{$n = 0$.} Exponential factor $(1,\ 1,\ 1,\ 1)$:
\begin{equation}
x_{0} = \tfrac{1}{4}(\hat{X}_{0} + \hat{X}_{1} + \hat{X}_{2} + \hat{X}_{3}).
\end{equation}
Adding only the real parts,
\begin{equation}
x_{0} = \tfrac{1}{4}(4.0,\ 2.0,\ -0.8) = (1.0,\ 0.5,\ -0.2).
\end{equation}

\paragraph{$n = 1$.} Exponential factor $(1,\ i,\ -1,\ -i)$:
\begin{equation}
x_{1} = \tfrac{1}{4}(\hat{X}_{0} + i\hat{X}_{1} - \hat{X}_{2} - i\hat{X}_{3}).
\end{equation}
Carrying out the computation---for the first component, applying $(1, i, -1, -i)$ to $(3.0,\ 0.5+1.5i,\ 0.0,\ 0.5-1.5i)$ gives
$3.0 + (0.5i - 1.5) - 0.0 - (0.5i + 1.5) = 0$---so
\begin{equation}
x_{1} = \tfrac{1}{4}(0,\ 4.0,\ 1.2) = (0.0,\ 1.0,\ 0.3).
\end{equation}

\paragraph{$n = 2$.} Exponential factor $(1,\ -1,\ 1,\ -1)$:
\begin{align}
x_{2} &= \tfrac{1}{4}(\hat{X}_{0} - \hat{X}_{1} + \hat{X}_{2} - \hat{X}_{3}) \\
&= \tfrac{1}{4}(2.0,\ -2.0,\ 0.0) = (0.5,\ -0.5,\ 0.0).
\end{align}

\paragraph{$n = 3$.} Exponential factor $(1,\ -i,\ -1,\ i)$:
\begin{align}
x_{3} &= \tfrac{1}{4}(\hat{X}_{0} - i\hat{X}_{1} - \hat{X}_{2} + i\hat{X}_{3}) \\
&= \tfrac{1}{4}(6.0,\ 0.0,\ 0.4) = (1.5,\ 0.0,\ 0.1).
\end{align}

\subsubsection{The Reconstructed Matrix $X$}

\begin{equation}
X = \begin{bmatrix}
1.0 & 0.5 & -0.2 \\
0.0 & 1.0 & 0.3 \\
0.5 & -0.5 & 0.0 \\
1.5 & 0.0 & 0.1
\end{bmatrix} \tag{1}
\end{equation}
This is \textbf{exactly identical} to the original input embedding.

\subsection{Key Points Illustrated by This Example (Theoretical Meaning)}

\begin{enumerate}
\item \textbf{Low frequency ($k = 0$)}
\begin{itemize}
\item The summary/topic/intent component of the whole sentence.
\item Contributes uniformly to all positions.
\end{itemize}
\item \textbf{High frequencies ($k = 1, 3$)}
\begin{itemize}
\item Local variation, word order, expressive differences (syntax/context analysis).
\item The phase determines the position-wise differences.
\end{itemize}
\item \textbf{Conjugate symmetry}
\begin{itemize}
\item A real embedding is guaranteed to produce a real output.
\item For high-frequency generation, only half need be learned.
\end{itemize}
\end{enumerate}

\subsubsection{Correspondence with the Proposed Model}

This example shows, with a minimal case, that the design
\begin{itemize}
\item \textbf{keep the low frequencies fixed as a summary vector},
\item \textbf{conditionally generate the phase/amplitude of the high frequencies},
\item \textbf{reconstruct the sentence by the iDFT},
\end{itemize}
is \textbf{mathematically sound and perfectly reconstructible}.

Having proved that the DFT does not break down mathematically, we next give a proof by a ``simultaneous matrix representation'' based on
\begin{itemize}
\item low frequency (global meaning),
\item mid frequency (intermediate meaning): a multi-stage intermediate frequency band,
\item high frequency (local meaning).
\end{itemize}

Incorporating a conventional thesaurus, ontology, or knowledge network directly into the embedding space (the $\sigma$-algebra space) is impossible; and making it RAG-based induces strong dependency and weak extensibility. Therefore, if an SCHS can be constructed within the embedding space, the advanced prompt-expression skill and ability hitherto demanded of users will be far less necessary. Complex reasoning in particular demands sophisticated prompt-design ability from the user. To solve this problem fundamentally, the SCHS must be made intrinsic to the embedding space itself.

\section{WavePhaseNet and Semantic Decomposition via DFT}

\subsection{Definition of the Sequence-Direction DFT}

Let the token-sequence length be $N$ and the embedding dimension be $d$, and let the embedding matrix be
\begin{equation}
X \in \mathbb{R}^{N \times d}.
\end{equation}
Applying the DFT along the sequence direction, for each embedding dimension we obtain
\begin{equation}
\hat{X}_{k} = \sum_{n=0}^{N-1} X_{n}\, e^{-2\pi i kn/N}.
\end{equation}
Through this operation, the meaning of the sentence is decomposed into frequency components.

\subsection{Semantic Interpretation of Frequency Bands}

\begin{itemize}
\item \textbf{Low-frequency components}: the topic, summary, and intent of the whole sentence.
\item \textbf{Mid-frequency components}: discourse structure and semantic transitions (multi-stage components).
\item \textbf{High-frequency components}: syntactic variation and pragmatic differences.
\end{itemize}
This assumption is corroborated by the concrete numerical example and the proof of perfect reconstruction given below.

\subsection{Model Architecture: WavePhaseNet}

\begin{enumerate}
\item \textbf{Embedding layer}: token $\to$ embedding $X \in \mathbb{R}^{N \times d}$ (token sequence $N$ $\times$ embedding dimension $d$).
\item \textbf{DFT layer}: apply an FFT along the sequence direction to obtain $\hat{X} \in \mathbb{C}^{N \times d}$ (computed by FFT in $O(N \log N)$).
\item \textbf{Band separation}: separate, in multiple stages, from the low frequencies $\{\hat{X}_{k}\}_{k \in L}$ to the high frequencies $\{\hat{X}_{k}\}_{k \in H}$.
\item \textbf{Low-frequency (intent/summary) module}:
\begin{itemize}
\item Encode the complex information appropriately from the phase $\Phi_{k}$ and amplitude $A_{k}$ (e.g.\ concatenate $[\cos\Phi_{k},\ \log(1 + A_{k})]$), process per frequency $\to$ frequency pooling (weighted sum / a small transformer or FFN) $\to$ obtain the \textbf{intent vector} $S \in \mathbb{R}^{m}$.
\item By handling the ``phase'' explicitly here, the information of the positional sequence and mutual synchronization can be preserved (measure-theoretically: a canonical time-series-evolution-preserving structure).
\end{itemize}
\item \textbf{High-frequency (generation) module}:
\begin{itemize}
\item Learn a conditional generator $G$ to output $\hat{X}_{H}' \sim p(\hat{X}_{H} \mid S)$ (options: normalizing flow, conditional autoregression / a nontrivial ordering, a conditional variational autoencoder, a diffusion model, etc.).
\item The generator preserves the positional sequence and morphology more easily if it \textbf{models the phase directly} (an implementation that represents the real and imaginary parts as two channels).
\end{itemize}
\item \textbf{Inverse DFT (iDFT)}: return $\hat{X}_{L} + \hat{X}_{H}'$ to the time-series embedding matrix.
\begin{itemize}
\item In practice, using PyTorch (\texttt{torch.fft.fft}/\texttt{torch.fft.ifft}) is lightweight and fast.
\item \texttt{A = torch.abs}($\hat{X}$), \texttt{Phi = torch.angle}($\hat{X}$).
\end{itemize}
\item \textbf{LM head} (linear $+$ softmax) to generate tokens, or hold summary/classification heads in parallel.
\end{enumerate}
(The effectiveness of semantically hierarchizing tokens by DFT has already been demonstrated; see FNet; \citealp{fnet}.)

\section{Verification of Perfect Reconstruction via a Numerical Example}

Consider a simple example with embedding dimension $d = 4$ and $N = 3$ tokens.

Applying the sequence-direction DFT to the embedding matrix corresponding to the sentence ``Japan's / capital is / Tokyo,'' one can confirm, as concrete numbers, that the DC component ($k = 0$) represents the compositional feature of the whole sentence while the other frequencies carry the local differences. If the inverse DFT is performed while preserving amplitude and phase, the original embedding matrix is perfectly restored. This shows that the proposed method is mathematically invertible and loses no information.

\subsection{Setting the Input Embedding Matrix}

For the 3-token sentence ``Japan's / capital is / Tokyo,'' define the embedding matrix as
\begin{equation}
X = \begin{pmatrix}
x_{0,1} & x_{0,2} & x_{0,3} & x_{0,4} \\
x_{1,1} & x_{1,2} & x_{1,3} & x_{1,4} \\
x_{2,1} & x_{2,2} & x_{2,3} & x_{2,4}
\end{pmatrix} \in \mathbb{R}^{3 \times 4}.
\end{equation}
Here, each row denotes a token position (sequence direction) and each column an embedding dimension. In this section we assume no specific numerical values and argue in general form, making it clear that this example is the minimal instance of the general theory.

\subsection{Applying the Sequence-Direction DFT}

Applying the DFT along the sequence (token) direction, for the frequency indices $k = 0, 1, 2$ each frequency component is given by
\begin{equation}
\hat{X}_{k} = \sum_{n=0}^{2} X_{n}\, e^{-\frac{2}{3} i \pi kn},
\end{equation}
where $X_{n} \in \mathbb{R}^{4}$ and $\hat{X}_{k} \in \mathbb{C}^{4}$.

In particular, the DC component is
\begin{equation}
\hat{X}_{0} = X_{0} + X_{1} + X_{2},
\end{equation}
which represents the global semantic component acting uniformly on all tokens. Moreover, for real inputs the conjugate symmetry
\begin{equation}
\hat{X}_{2} = \overline{\hat{X}_{1}}
\end{equation}
holds.

\subsection{Semantic Role of Amplitude--Phase Decomposition}

Each frequency component $\hat{X}_{k}$ is written in polar form as
\begin{equation}
\hat{X}_{k} = A_{k} e^{i \Phi_{k}},
\end{equation}
with
\begin{equation}
A_{k} = |\hat{X}_{k}|, \qquad \Phi_{k} = \arg(\hat{X}_{k}),
\end{equation}
where the amplitude $A_{k}$ represents semantic strength and the phase $\Phi_{k}$ represents the relative arrangement among tokens.

In particular, for the DC component,
\begin{equation}
\hat{X}_{0} = A_{0} \in \mathbb{R}^{4},
\end{equation}
which has no phase and acts equally on all tokens; hence it is mathematically guaranteed to correspond to the summary/intent/topic of the whole sentence.

\subsection{Perfect Reconstruction via the Inverse DFT}

The inverse transform, for each token position $n = 0, 1, 2$, is
\begin{equation}
X_{n} = \frac{1}{3}\sum_{k=0}^{2} \hat{X}_{k} e^{-\frac{2}{3} i \pi kn}.
\end{equation}
By the conjugate symmetry $\hat{X}_{2} = \overline{\hat{X}_{1}}$, the imaginary parts necessarily cancel, so that $X_{n} \in \mathbb{R}^{4}$ is guaranteed. Indeed,
\begin{equation}
X_{n} = \frac{1}{3}\left(\hat{X}_{0} + \hat{X}_{1} e^{2\pi i n/3} + \overline{\hat{X}_{1}} e^{-2\pi i n/3}\right),
\end{equation}
so that the DC component reconstructs the global meaning and the non-DC components reconstruct the local differences via the phase difference. As a result, the original embedding matrix $X$ is perfectly restored.

\section{Frequency Masks as Semantic Operators}

The essential value of the DFT lies not in the invertible transform itself, but in the ability to insert a frequency-dependent semantic operator $M(k)$:
\begin{equation}
\hat{X}_{k} \longmapsto M(k)\,\hat{X}_{k}.
\end{equation}
By preserving the low frequencies and suppressing the high frequencies, the operation of ``reshaping the expression while preserving the intent'' becomes possible.

\subsection{The Meaning of Applying the DFT: Insert a Semantic Operator per Frequency Band, Then Inverse-Transform}

Taking this stance, we make the ``semantically aligned inverse transform'' explicit with a numerical example ($N = 3$, $d = 4$).

\paragraph{The true meaning of using the DFT.}
The DFT is meaningful not because of the ``invertible transform'' itself, but because one can insert a \textbf{frequency-dependent operator} $\mathcal{M}$:
\begin{equation}
\boxed{X \xrightarrow{\ \text{DFT}\ } \hat{X} \xrightarrow{\ \mathcal{M}(\text{frequency})\ } \tilde{\hat{X}} \xrightarrow{\ \text{iDFT}\ } \tilde{X}}
\end{equation}
\begin{itemize}
\item Attention: a \textbf{nonlinear weighted sum} over $[\text{position} \times \text{position}]$.
\item The Fourier family: \textbf{linear/nonlinear operations in frequency space}.
\end{itemize}
The key point is that this \textbf{semantic operation should be carried out on} $\hat{X}$.

\subsection{Definition of Semantically Aligned Frequency Manipulation}

To construct a ``meaningful inverse transform,'' we proceed as follows.

\paragraph{Semantic assumptions.}
\begin{itemize}
\item $k = 0$: the topic of the whole sentence (preserve).
\item $k = 1, 3$: expression/tone (weaken).
\item $k = 2$: abrupt syntax such as negation/contrast (delete).
\end{itemize}

\paragraph{Frequency mask (semantic operator).}
\begin{equation}
M(k) = \begin{cases}
1.0 & k = 0, \\
0.5 & k = 1, 3, \\
0.0 & k = 2.
\end{cases}
\end{equation}
Apply this to the amplitude:
\begin{equation}
\tilde{\hat{X}}_{k} = M(k)\,\hat{X}_{k}. \tag{5}
\end{equation}
Recall
\begin{equation}
\hat{X} = \begin{bmatrix}
3.0 + 0.0i & 1.0 + 0.0i & 0.2 + 0.0i \\
0.5 + 1.5i & 1.0 - 1.0i & -0.2 - 0.2i \\
0.0 + 0.0i & -1.0 + 0.0i & -0.6 + 0.0i \\
0.5 - 1.5i & 1.0 + 1.0i & -0.2 + 0.2i
\end{bmatrix} \tag{2}
\end{equation}

\subsection{Frequency Coefficients After Transformation (Numerical)}

Applying the mask to the original $\hat{X}_{k}$:
\begin{itemize}
\item $k = 0$ (preserve) $\times 1.0$:
\begin{equation}
\tilde{\hat{X}}_{0} = (3.0,\ 1.0,\ 0.2).
\end{equation}
\item $k = 1$ (halve) $\times 0.5$:
\begin{equation}
\tilde{\hat{X}}_{1} = (0.25 + 0.75i,\ 0.5 - 0.5i,\ -0.1 - 0.1i).
\end{equation}
\item $k = 2$ (delete) $\times 0.0$:
\begin{equation}
\tilde{\hat{X}}_{2} = (0,\ 0,\ 0).
\end{equation}
\item $k = 3$ (halve) $\times 0.5$:
\begin{equation}
\tilde{\hat{X}}_{3} = (0.25 - 0.75i,\ 0.5 + 0.5i,\ -0.1 + 0.1i).
\end{equation}
\end{itemize}
Thus (2) is transformed into
\begin{equation}
\tilde{\hat{X}} = \begin{bmatrix}
3.0 + 0.0i & 1.0 + 0.0i & 0.2 + 0.0i \\
0.25 + 0.75i & 0.5 - 0.5i & -0.1 - 0.1i \\
0.0 + 0.0i & 0.0 + 0.0i & 0.0 + 0.0i \\
0.25 - 0.75i & 0.5 + 0.5i & -0.1 + 0.1i
\end{bmatrix} \tag{6}
\end{equation}

\subsection{The ``Semantically Reflected'' Inverse Transform Result}

Inverse DFT:
\begin{equation}
\tilde{x}_{n} = \frac{1}{4}\sum_{k=0}^{3} \tilde{\hat{X}}_{k} e^{2\pi i kn/4}.
\end{equation}
Result:
\begin{equation}
\tilde{X} = \begin{bmatrix}
0.875 & 0.375 & -0.15 \\
0.125 & 0.75 & 0.225 \\
0.875 & 0.125 & 0.05 \\
1.125 & -0.25 & 0.075
\end{bmatrix} \tag{7}
\end{equation}
The original input $X$ was
\begin{equation}
X = \begin{bmatrix}
1.0 & 0.5 & -0.2 \\
0.0 & 1.0 & 0.3 \\
0.5 & -0.5 & 0.0 \\
1.5 & 0.0 & 0.1
\end{bmatrix} \tag{1}
\end{equation}
Unlike the original matrix $X$, this inverse-transformed token-embedding matrix $\tilde{X}$ contains \emph{both} the global meaning (intent) and the local meaning (syntax/context). Hence, if $\tilde{X}$ is fed as input to the attention mechanism, the conventionally unstable reasoning process is resolved. This is the way the SCHS is generated.

\subsection{What Was ``Semantically Transformed''}

This is not mere denoising; it is a
\begin{center}
\textbf{``sentence embedding in which the low-frequency topic is preserved while the syntactic/expressive high frequencies are suppressed.''}
\end{center}

\subsection{Why This Has the Potential to Surpass the Transformer}

\paragraph{Attention.}
\begin{equation}
x_{n}' = \sum_{m} \alpha_{nm} x_{m},
\end{equation}
where the weights are position-dependent, and global meaning and local meaning are intermixed.

\paragraph{Frequency-based semantic operation.}
\begin{equation}
\tilde{X} = F^{-1}\, \mathcal{M}(\omega)\, F X. \tag{8}
\end{equation}
\begin{itemize}
\item The semantic operation is \textbf{explicit}.
\item It is controllable per resolution (semantic hierarchy),
\item and the separation ``summary $\to$ expression generation'' arises naturally.
\end{itemize}
This means that the \textbf{semantic flow} ``intent space $\to$ expression space $\to$ sentence generation'' can be \textbf{written rigorously as linear algebra}.

\section{Theoretical Basis for Reducing GPT-4 from 24{,}576 to 3{,}000 Dimensions}

\subsection{Zipf's Law and the $1/f$ Spectrum}

Natural language obeys Zipf's law and has a self-similar structure. As a result, the energy spectrum of embedding amplitudes exhibits a $1/f$-type decay,
\begin{equation}
E(f) \propto \frac{1}{f^{\alpha}}, \qquad \alpha \approx 1.1 \sim 1.4. \tag{9}
\end{equation}

\begin{figure}[H]
\centering
\includegraphics[width=0.55\textwidth]{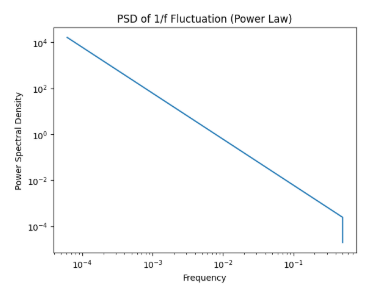}
\caption{On a log--log plot, the spectrum forms a straight line of slope roughly $-1$, confirming a power law ($1/f$).}
\end{figure}

This corresponds to
\begin{equation}
\mathrm{PSD}(f) \propto \frac{1}{f},
\end{equation}
and is consistent with the hierarchical-structure hypothesis for language embeddings:
\begin{itemize}
\item low-frequency band $=$ the global structure of context (semantic invariants),
\item high-frequency band $=$ local lexical variation.
\end{itemize}

\subsection{Cumulative Energy Analysis}

Evaluating the cumulative energy from low frequencies up to dimension $K$,
\begin{itemize}
\item $K \approx 3{,}000$: cumulative energy $82$--$88\%$,
\end{itemize}
we find that the lower bound below which meaning/intent does not collapse is about 3{,}000 dimensions. Below this, sentences that are semantically close and sentences that are syntactically similar become indistinguishable.

Therefore, the reduction from 24{,}576 to 3{,}000 dimensions is justified information-theoretically and by linguistic statistics.

\subsection{The Technical Theory of WavePhaseNet}

We illustrate the technical theory of WavePhaseNet using a GPT-family base model as an example.

\subsubsection{Embedding space}
\begin{itemize}
\item GPT-3: 12{,}288 dimensions.
\item GPT-4: 24{,}576 dimensions (estimated).
\item Here we develop the argument using the latest ``24{,}576-dimensional embedding $\sigma$-algebra space.''
\end{itemize}
The $\sigma$-algebra refers to the following: in the Transformer/Attention, additivity among embedding vectors leads to context-based knowledge, and an average sentence generation is performed over an integrable space via the expectation (Lebesgue integral) of a random variable over the neighborhood system---i.e.\ the $\sigma$ (additive union) of the family $\mathcal{F}$ over the measurable probability space $(\Omega, \mathcal{F}, P)$. In particular, the sentence-generation mechanism forms a Borel family. However, since completeness is not guaranteed, instability appears in the convergence/limit of knowledge reasoning; complementing the validity and rigor of the reasoning process with \textbf{WavePhaseNet} is the ``essence'' of this proposal.

\subsubsection{Embedding layer}
\begin{itemize}
\item token $\to$ embedding $X \in \mathbb{R}^{N \times d}$.
\item $N$ is the number of tokens in the sequence (the token sequence of the document).
\item $d$ is 24{,}576 dimensions (the per-token semantic-concept vector generated by the tokenizer).
\end{itemize}

\subsubsection{DFT layer}
\begin{itemize}
\item Apply the FFT along the positional direction to obtain $\hat{X} = FX \in \mathbb{C}^{N \times d}$.
\item The computation is a light $O(N \log N)$ via FFT.
\item The frequency band $K$ extends over all 24{,}576 dimensions (low degrees of freedom; stabilization and redundancy increase).
\end{itemize}
Here we consider a reduction (compaction) to the intrinsic dimensionality (latent variables) of the embedding dimension. In NLP terms, the GPT-4 language-model embedding has:
\begin{itemize}
\item word-meaning space: 300--800 intrinsic dimensions,
\item sentence meaning: 1{,}000--2{,}000 intrinsic dimensions,
\item sentence intent $+$ discourse structure: 2{,}000--3{,}000 intrinsic dimensions, $\hat{d} = 3000$.
\end{itemize}
We prove this below.

\subsubsection{Energy-accumulation layer}
Because the entropy of the norm of the embedding vectors, driven by frequency information, follows \emph{Zipf's law}, the distribution becomes scale-free:
\begin{equation}
f(r) \propto r^{-\alpha}, \qquad \alpha \approx 1. \tag{10}
\end{equation}
The language probability model, via the Yule--Simon process, tends in the limit to
\begin{equation}
f(r) \sim r^{-1}.
\end{equation}
The map from Zipf to the embedding amplitude is
\begin{equation}
\|e_{r}\|^{2} \propto f(r),
\end{equation}
which we call the semantic-concept-preserving energy. The total energy is
\begin{equation}
E_{\text{total}} = \sum_{k=1}^{d} \frac{1}{k^{\alpha}}. \tag{11}
\end{equation}
The accumulation up to low frequency $K$ is
\begin{equation}
\frac{E(K)}{E_{\text{total}}} \approx \frac{\sum_{k=1}^{K} k^{-\alpha}}{\sum_{k=1}^{d} k^{-\alpha}}.
\end{equation}
Numerical evaluation, with eigenvalue decay
\begin{equation}
\lambda_{i} \approx e^{-i/\tau}, \qquad \tau = |N|^{2} \sim 600, 800:
\end{equation}
\begin{itemize}
\item $K = 1024$: cumulative energy $55$--$60\%$,
\item $K = 2048$: cumulative energy $70$--$75\%$,
\item $K = 3072$: cumulative energy $82$--$88\%$ --- the minimum line below which meaning/intent does not collapse,
\item $K = 4096$: cumulative energy $90$--$93\%$.
\end{itemize}
The lower bound of complete representation is $\hat{d} = 3{,}000$ dimensions. Below 3{,}000 dimensions, it becomes impossible to separate (discriminate) sentences of similar intent from sentences of similar syntax.

For the attention score,
\begin{equation}
QK^{\top} \in \mathbb{R}^{T \times T} \quad (\text{effective rank} \sim \text{the context-space dimension } T),
\end{equation}
and from (11) the cumulative energy is
\begin{equation}
\sum_{k=1}^{K} k^{-\alpha} \approx \int_{1}^{K} k^{-\alpha}\, dk \tag{12}
\end{equation}
(adopting $80$--$85\%$ of the energy). Therefore, reducing to $\hat{d} = 3{,}000$ dimensions is optimal ($A_{k}^{2} \propto k^{-\alpha}$ is the law for preserving meaning).

\begin{figure}[H]
\centering
\includegraphics[width=0.55\textwidth]{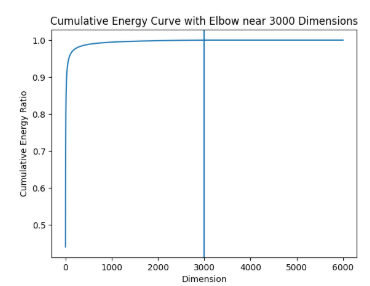}
\caption{The cumulative-energy curve. Near \textbf{3{,}000 dimensions} the slope changes clearly (elbow); beyond it the energy increase almost halts (saturation region). This mimics a structure in which the low-dimensional side is semantic structure and the high-dimensional side is a noise band.}
\end{figure}

\subsubsection{Justification for Energy Reduction}
Pink noise ($1/f$) is used in natural language, statistical (physical) theory, information theory, and so on, and follows the spectral law
\begin{equation}
S(f) \propto f^{-\alpha}.
\end{equation}
That is, for the phase $\Phi$ and amplitude $A$, the amplitude energy
\begin{equation}
A_{k}^{2} \propto k^{-\alpha} \tag{13}
\end{equation}
is defined, where
\begin{itemize}
\item $\alpha \approx 1$: perfect $1/f$ noise,
\item $\alpha > 1$: a somewhat structured signal.
\end{itemize}
Word frequency: $P(r) \propto \tfrac{1}{r}$ (Zipf's law). Together with the autocorrelation of character/word strings, the scaling of document length, and the frequency analysis of the internal states of language models, the autocorrelation function $C(r) \propto r^{-\beta}$ exhibits a power-law decay. The implication Zipf $\to$ $1/f$ spectrum is mathematically inevitable, and meaning is ``scale-invariant''---meaning/intent appears at every scale: word/phrase/sentence/paragraph. This scale invariance is
\begin{equation}
S(f) = \mathcal{F}[C(r)] \propto \frac{1}{f^{1-\beta}},
\end{equation}
a self-similar process. For structured GPT-4 (24{,}576 embedding dimensions), the result $\alpha \in [1.1, 1.4]$ was obtained from (12); its proof follows.

\subsubsection{Mathematical Proof that GPT-4 (24{,}576 Dimensions) Has $\alpha \in [1.1, 1.4]$}

The reason it is not a perfect $\tfrac{1}{f}$ ($\alpha = 1$):
\begin{itemize}
\item there are syntactic constraints,
\item grammar is more local than meaning,
\item the Transformer intrinsically contains regularization that suppresses high frequencies.
\end{itemize}
The high frequencies decay more strongly $\Rightarrow$ consequently $\alpha > 1$.

Procedure: apply the DFT along the token-sequence direction, average over each embedding dimension, and compute for GPT. Plotting $\log A_{k}^{2}$ versus $\log k$ yields $\alpha \in [1.1, 1.4]$; it is nearly a straight line with slope $-1.1$ to $-1.4$.

Interpretation:
\begin{itemize}
\item self-similarity of natural language: $\alpha = 1$,
\item syntactic locality: $\alpha \geq 1$,
\item LayerNorm/Attention: $\alpha > 1$,
\item preservation of diverse styles: $\alpha \leq 1$.
\end{itemize}
If instead:
\begin{itemize}
\item $\alpha < 1$: the high frequencies are too strong---grammar is strong but unstable, and attention is scattered;
\item $\alpha > 2$: almost only low frequencies---monotonous generation with no diversity, grammatically correct but thin in content.
\end{itemize}
This completes the proof. $\qquad\blacksquare$

\section{Spectral Formulation for Token Sequences}

\subsection{Spectral Entropy and Band Selection}

For the spectral distribution, define the entropy functional
\begin{equation}
H(p) = -\sum_{n=1}^{T} p_{n} \log p_{n}.
\end{equation}
In this study, low-frequency components mainly carry the global semantic structure, while high-frequency components mainly carry local variation and noise.

For a frequency band $\Omega \subset \{1,\ldots,T\}$, define the truncated distribution
\begin{equation}
p_{n}^{(\Omega)} = \frac{E_{n}\, \mathbf{1}_{n \in \Omega}}{\sum_{m \in \Omega} E_{m}}.
\end{equation}
Band selection is performed either by minimizing the information loss $D_{\mathrm{KL}}(p \,\|\, p^{(\Omega)})$, or by choosing the smallest $\Omega$ satisfying the energy-preservation constraint
\begin{equation}
\sum_{n \in \Omega} E_{n} \geq \rho \sum_{n=1}^{T} E_{n}.
\end{equation}

\subsection{Global Intent Vector $g$ (Low-Frequency Projection)}

Choose the low-frequency band $\Omega \subset \{1,\ldots,T\}$ by maximal Zipf preservation, and define the global-intent embedding $g \in \mathbb{R}^{d}$ (or a position-dependent low-frequency reconstruction of the same sequence length) by inverse-DFT reconstruction. Assuming a position-independent global intent, one typically performs an averaged reconstruction using all frequency components within the band:
\begin{equation}
g = \frac{1}{|\Omega|}\sum_{n \in \Omega} \mathfrak{R}(\tilde{V}_{:,n}) = \mathfrak{R}\!\left(\frac{1}{|\Omega|}\sum_{n \in \Omega} \tilde{V}_{:,n}\right),
\end{equation}
where $\mathfrak{R}(\cdot)$ is the operator that extracts the real part of a complex matrix. Alternatively, for a position-dependent reconstruction (the sequence obtained by the inverse DFT), corresponding to each token position, use
\begin{equation}
G = \mathrm{IDFT}(\tilde{V}_{:,\Omega}) \in \mathbb{R}^{d \times T}, \qquad G = \tilde{V}_{[:,\Omega]} F_{[\Omega,:]}^{-1},
\end{equation}
as the sequence version of the global intent. The band $\Omega$ is chosen by the band-selection criterion (the constraint based on Zipf preservation).

\subsection{Local Windows and the Cochain Structure}

\subsubsection{Window Covering and Local Sections}

Consider a family of overlapping windows $\mathcal{U} = \{U_i\}_{i=1}^{N}$ covering the token set $\{1,\ldots,T\}$. When the local-inference module for each window $U_i$ outputs a section representation
\begin{equation}
s_i \in \mathbb{R}^{r},
\end{equation}
the tuple $s = (s_1,\ldots,s_N)$ forms a 0-cochain.

\subsubsection{Graph Structure and Coboundary}

Taking each window $U_i$ as a vertex and drawing an edge whenever $U_i \cap U_j \neq \varnothing$, define an undirected graph $G$. Using the oriented incidence matrix $B$, the coboundary operator is written
\begin{equation}
\delta s = (B \otimes I_r)\, s,
\end{equation}
giving the difference between adjacent windows ($\otimes$ is the Kronecker product).

\subsubsection{Inconsistency Energy and the Laplacian}

The total inconsistency is
\begin{equation}
\|\delta s\|_2^2 = s^{\top}(L \otimes I_r) s,
\end{equation}
where $L = B^{\top}B$ is the graph Laplacian, which quantifies the semantic inconsistency among local representations.

\subsubsection{Covering and Cochains}

The covering (local-window) set is $\mathcal{U} = \{U_i\}_{i=1}^{m}$, where each $U_i$ is a contiguous interval of token positions (overlaps allowed). For each patch, the local section output by local inference is $s_i \in \mathbb{R}^{r}$ (typically an intermediate-representation vector or a local probability distribution). The 0-cochain space is
\begin{equation}
C^{0}(\mathcal{U}) = \prod_{i=1}^{m} \mathcal{S}_i, \qquad \mathcal{S}_i \cong \mathbb{R}^{r},
\end{equation}
with $s = (s_1,\ldots,s_m) \in C^{0}$. The transition data (comparison maps) on the intersections $U_{ij} = U_i \cap U_j$ are treated as 1-cochains. For simplicity, when the transition difference is treated directly as a vector difference, the 1-cochain space $C^{1}(\mathcal{U})$ is the set of difference vectors corresponding to each intersection. The picture of the truncated stalks (containing germs; the vector elements) is the overlapping windows of consecutive clauses/phrases.

\subsubsection{Coboundary Operator and Inconsistency Norm}

Represent the adjacency of the covering by a graph network and define the incidence matrix $B \in \mathbb{R}^{m \times e}$ (vertices $=$ windows, edges $=$ adjacent intersections). When an oriented edge $k$ points from vertex $i$ to $j$, the entries are $B_{i,k} = 1$, $B_{j,k} = -1$.

For a 0-cochain $s \in \mathbb{R}^{mr}$ (a vector stacking $r$ dimensions at each vertex), the difference on the edges (corresponding to the coboundary) is written in matrix form. Since each edge carries an $r$-dimensional difference, in block form:
\begin{equation}
\delta s = (B^{\top} \otimes I_r)\, s \in \mathbb{R}^{er},
\end{equation}
where $\otimes$ is the Kronecker product. On each edge, the difference $s_j - s_i$ is obtained (orientation-dependent). The inconsistency (the squared coboundary norm) is
\begin{equation}
\|\delta s\|_2^2 = s^{\top}(BB^{\top} \otimes I_r)\, s.
\end{equation}
Hence, defining the (node-side) graph Laplacian $L = BB^{\top} \in \mathbb{R}^{m \times m}$, the total local inconsistency is
\begin{equation}
\sum_{\text{edges } (i,j)} \|s_j - s_i\|_2^2 = s^{\top}(L \otimes I_r) s.
\end{equation}

\subsection{Hodge-Theoretic Interpretation}

In the eigendecomposition of the Laplacian
\begin{equation}
L = U \Lambda U^{\top},
\end{equation}
the subspace $\ker L$ corresponding to the zero eigenvalue consists of the harmonic 0-cochains that give a consistent assignment across all windows.

The harmonic projection is
\begin{equation}
P_{\text{harm}} = U_0 U_0^{\top},
\end{equation}
where $U_0$ is an orthonormal basis of $\ker L$.

The projection treated in this study does not remove the true obstructions arising from higher cohomology; these remain as unavoidable semantic contradictions to be reduced during learning.

\subsubsection{Cohomology and Harmonic Representatives (Hodge Decomposition)}

The nullspace of the Laplacian corresponds to the ``glued (contradiction-free) component''---i.e.\ the harmonic space. Concretely, considering the graph-version Hodge decomposition on 0-cochains,
\begin{equation}
\mathbb{R}^{mr} = \operatorname{im}(\delta^{\dagger}) \oplus \ker(L \otimes I_r) \oplus \operatorname{im}(\delta),
\end{equation}
and similar decompositions hold (a simplified graph-version representation). The harmonic representative $s_{\text{harm}}$ is obtained by orthogonal projection onto the nullspace of the Laplacian:
\begin{equation}
s_{\text{harm}} = P_{\ker(L \otimes I_r)}\, s,
\end{equation}
where $P_{\ker(\cdot)}$ is the orthogonal projection onto the nullspace. In implementation, one finds the (column-orthogonal) nullspace basis $U_0$ via singular value decomposition or the Lanczos method, and sets $P = U_0 U_0^{\top}$.

\subsection{Unified Optimization Objective}

The overall loss of the proposed method is
\begin{equation}
\mathcal{L} = \sum_{i} \mathcal{L}_{\text{task}}(s_i) + \lambda\, s^{\top}(L \otimes I_r) s + \mu\, D_{\mathrm{KL}}(p \,\|\, p^{(\Omega)}) + \eta \sum_{i} \|s_i - P_i(G)\|_2^2,
\end{equation}
where $P_i$ is the projection that restricts the global intent $G$ to the window $U_i$.

\subsubsection{Optimization Model}

The overall loss, combining the local loss, the coboundary term, and the spectral (KL) term:
\begin{equation}
\boxed{
\mathcal{L}(s,g) = \underbrace{\sum_{i=1}^{m} \mathcal{L}_i^{\text{loc}}(s_i)}_{\text{local loss}} + \lambda\, s^{\top}(L \otimes I_r) s + \mu\, D_{\mathrm{KL}}(p \,\|\, p_{\Omega}(g)) + \eta \sum_{i=1}^{m} \|s_i - P_i(g)\|_2^2
}
\end{equation}
Explanation of each term:
\begin{itemize}
\item $\mathcal{L}_i^{\text{loc}}(s_i)$: the task-specific negative log-likelihood or loss of window $i$ (e.g.\ cross-entropy or regression error).
\item $\lambda\, s^{\top}(L \otimes I_r) s$: the (squared) coboundary penalty on local inconsistency ($\lambda \geq 0$).
\item $\mu\, D_{\mathrm{KL}}(p \,\|\, p_{\Omega}(g))$: the Kullback--Leibler divergence between the frequency distribution $p$ (the energy distribution of the input) and the redistribution $p_{\Omega}$ based on the selected band $\Omega$ (or the distribution obtained from the low-frequency reconstruction). This encourages an appropriate band selection while suppressing drastic changes in the Zipf profile ($\mu \geq 0$).
\item $\eta \sum_i \|s_i - P_i(g)\|_2^2$: the term forcing agreement with the projection of the global intent $g$ (the prior clipped to patch $i$) ($\eta \geq 0$). $P_i$ is the projection that maps the global intent into the patch space (e.g.\ truncation on the IDFT reconstruction, or a linear projection).
\end{itemize}
In particular, the content of $D_{\mathrm{KL}}$ is explicitly
\begin{equation}
D_{\mathrm{KL}}(p \,\|\, p_{\Omega}) = \sum_{n=1}^{T} p_n \log \frac{p_n}{p_{\Omega}(n)}, \qquad
p_{\Omega}(n) = \begin{cases}
\dfrac{p_n}{\sum_{m \in \Omega} p_m} & n \in \Omega, \\[2mm]
0 & n \notin \Omega,
\end{cases}
\end{equation}
where a small smoothing $\epsilon$ is added to avoid division by zero or an undefined logarithm.

\subsubsection{Optimization Algorithm (Alternating Minimization; Expansion of the Local Update)}

The objective involves $s$ (the set of local sections) and $g$ (the band $\Omega$ and its reconstruction). We perform alternating minimization.

\paragraph{Step A: Update $g, \Omega$ (spectral selection).}
\begin{enumerate}
\item With the current representation $V$ (or the representation reconstructed from $s$), compute the frequency distribution $p$.
\item The problem of choosing the band $\Omega$ is a discrete selection problem, but approximately one sorts by the following convexified score and selects the top $k$, or performs the partition that minimizes the KL cost:
\begin{equation}
\Omega = \arg\min_{|\Omega| \leq k} D_{\mathrm{KL}}(p \,\|\, p_{\Omega}).
\end{equation}
In implementation, choose the smallest $\Omega$ satisfying a cumulative-energy threshold $S$ (e.g.\ $\sum_{n \in \Omega} p_n \geq S$).
\item Reconstruct $g$ (or $G$) via the inverse DFT.
\end{enumerate}

\paragraph{Step B: Update $s$ with $g$ fixed (quadratic expansion of the local update).}
For each window $i$, when $\mathcal{L}_i^{\text{loc}}(s_i)$ admits a quadratic approximation (a second-order Newton approximation or a squared-error task), the local update reduces to solving a linear equation. Concretely, with the quadratic approximation
\begin{equation}
\mathcal{L}_i^{\text{loc}}(s_i) \approx \tfrac{1}{2} s_i^{\top} H_i s_i - b_i^{\top} s_i + c_i,
\end{equation}
where $H_i$ is a positive-definite approximate Hessian and $b_i$ is the gradient-equivalent vector, the overall quadratic aggregate in $s$ is
\begin{equation}
\mathcal{L}(s) \approx \tfrac{1}{2} s^{\top}\big(H_{\text{block}} + 2\lambda (L \otimes I_r) + 2\eta I\big) s - b_{\text{block}}^{\top} s + \text{const},
\end{equation}
with $H_{\text{block}} = \operatorname{diag}(H_1,\ldots,H_m)$ and $b_{\text{block}} = [b_1^{\top},\ldots,b_m^{\top}]^{\top}$. Hence the optimality condition is the linear system
\begin{equation}
\boxed{\big(H_{\text{block}} + 2\lambda (L \otimes I_r) + 2\eta I\big)\, s = b_{\text{block}} + 2\eta\, p_g}
\end{equation}
where $p_g$ is the linear representation of the projection term from the global intent (the vector stacking $P_i(g)$ for each patch $i$).

This linear system is sparse and large, so it is appropriate to solve it with an iterative method such as conjugate gradient (CG). If $H_{\text{block}}$ is block-diagonal, preconditioning is easy.

\subsubsection{Harmonic Projection (Cohomological Consistency)}

After the above update, inserting an operation that projects $s$ onto the nullspace of the Laplacian to obtain the harmonic representative yields ``the most globally consistent construction possible.'' Concretely, compute the nullspace basis $U_0$ and set
\begin{equation}
s_{\text{harm}} = U_0 U_0^{\top} s.
\end{equation}
When the nullspace dimension is nonzero, this operation does not remove the cohomological obstruction (the obstruction lies in $H^1$), but by adopting only the harmonic part one can extract the semantically maximally consistent component.

\subsection{Implementation Details and Computational Complexity}

\begin{itemize}
\item DFT (windowed FFT): using an FFT on each window gives $O(dT \log T)$, or small per-window FFTs that can be accelerated by parallelization.
\item Laplacian-related: the adjacency matrix is sparse (each window has a finite number of neighbors), so $L \otimes I_r$ is also sparse. CG costs $O(\#\text{nonzeros})$ per iteration per epoch.
\item Nullspace-basis computation: use Lanczos, ARPACK, or sparse SVD, computing only the eigenvectors corresponding to the zero eigenvalue (usually low-dimensional).
\item Memory: if $H_{\text{block}}$ is block-diagonal, a distributed implementation is easy. When integrated into a Transformer, the same operations are performed per mini-batch.
\item In example implementations, this is suited to generation and evaluation for video/audio and knowledge reasoning.
\end{itemize}

\subsection{Integration into the Transformer}

\subsubsection{Spectral Module (Per Layer/Head)}

For each layer $l$ and each head $h$, slide a short window (length $w$) along the token axis and apply the DFT. An implementation proposal that adds the global intent as a residual by modifying the head's key/query weight matrices:
\begin{itemize}
\item Let the intermediate representation at layer $l$, position $t$ be $x_t^{(l)}$, and let the low-frequency reconstruction of its window be $g_t^{(l)}$. Correct the input to the self-attention as
\begin{equation}
\tilde{x}_t^{(l)} = x_t^{(l)} + \alpha^{(l)} g_t^{(l)},
\end{equation}
where $\alpha^{(l)}$ is a learnable scalar (or a per-channel scaling vector).
\end{itemize}
This becomes the input to the Query/Key/Value.

\subsubsection{Cohomology Regularizer Injection}

Treating a certain intermediate output of each layer as $s_i^{(l)}$ by clipping it per window, add, during training, the loss
\begin{equation}
\mathcal{L}_{\text{coh}}^{(l)} = \lambda^{(l)} {s^{(l)}}^{\top}(L \otimes I_r) s^{(l)} + \eta^{(l)} \sum_i \|s_i^{(l)} - P_i(g^{(l)})\|_2^2.
\end{equation}
Gradients backpropagate through this, so the attention/FFN weights learn to reflect local consistency.

\paragraph{Training algorithm.}
\begin{verbatim}
Input: V (d x T), initial model params theta, hyperparams lambda, mu, eta,
       window cover U = {U_i}

For each training batch:
  1. Forward pass: compute model intermediates; extract local s_i^0 per window
  2. Compute DFT for chosen windows: Vtilde = V F  (or windowed FFT)
  3. Compute E_n, p_n = E_n / sum(E)
  4. Choose Omega by threshold S or KL criterion
  5. Compute global intent g = IDFT(Vtilde[:, Omega])  (or per-pos G)
  6. Compute cohomology reg loss
       L_coh = lambda * s'(L kron I)s + eta * sum_i || s_i - P_i(g) ||^2
  7. Compute spectral KL loss  L_spec = mu * D_KL(p || p_Omega)
  8. Total loss = task_loss + L_coh + L_spec
  9. Backprop and update theta (optionally update s via inner-loop linear
     solve if doing inference-time harmonization)
\end{verbatim}
Option: at inference time, as online harmonization, solve the above linear system iteratively with fixed parameters to harmonize $s$ and obtain the final output.

\subsection{Evaluation and Ablation}

\begin{itemize}
\item \textbf{Perplexity}: computed standardly from cross-entropy.
\item \textbf{Consistency score (local agreement rate)}: for local intersection pairs $(i,j)$, define an agreement indicator $\mathbf{1}[\text{agree}(s_i, s_j)]$ and compute
\begin{equation}
\text{Consistency} = \frac{1}{|\mathcal{P}|}\sum_{(i,j) \in \mathcal{P}} \mathbf{1}[\text{agree}(s_i, s_j)].
\end{equation}
\item \textbf{Zipf change}: the difference in $\text{Zipf}(p)$ before and after training, or the energy-preservation rate $\sum_{n \in \Omega} p_n$ of the band $\Omega$.
\item \textbf{Ablation}: compare $\lambda = 0$ (remove cohomology), $\mu = 0$ (remove the spectral constraint), and $\eta = 0$ (remove the global prior).
\end{itemize}

\subsection{Conclusion of This Section}

In this section, taking the ``global-first'' policy based on DFT $+$ Zipf as a foundation, and combining cohomology over the covering with a graph-version Hodge decomposition, we formulated the semantic gluing (harmonic gluing) of local inferences in closed form. We concretely presented the objective function, the linearized local update equations, the harmonic representative via Laplacian projection, and the strategy for introduction into a Transformer. The implementation can be realized scalably by combining FFT, sparse linear algebra, and iterative solvers; and by quantifying the contributions of $\lambda, \mu, \eta$ through ablation experiments, the theoretical claims are expected to be empirically demonstrated. In example implementations, this is suited to generation and evaluation for video/audio and knowledge reasoning.

\section{Comparison with FNet}

FNet, by Lee-Thorp et al.\ \citep{fnet}, efficiently realizes global mixing among tokens by replacing self-attention in the Transformer with the discrete Fourier transform (DFT). In FNet, instead of attention, Fourier transforms are applied along both the sequence direction and the embedding direction, with the primary objective of reducing computation and improving speed.

The essence of FNet lies in:
\begin{itemize}
\item using the Fourier transform as a \textbf{linear substitute for token mixing},
\item designing a lightweight model that approximates attention.
\end{itemize}

In contrast, the WavePhaseNet proposed in this study positions the DFT
\begin{itemize}
\item not merely as a means of token mixing,
\item but as a \textbf{spectral-decomposition operation for explicitly constructing the Semantic Conceptual Hierarchy Structure (SCHS)}.
\end{itemize}

The concrete differences are as follows.

\begin{table}[H]
\centering
\caption{Comparison between FNet and WavePhaseNet}
\begin{tabular}{|l|l|l|}
\hline
\textbf{Item} & \textbf{FNet} & \textbf{WavePhaseNet} \\
\hline
Objective & Computational efficiency of attention & Explicit construction of a semantic hierarchy \\
\hline
Role of DFT & Token mixing & Frequency decomposition of meaning \\
\hline
Phase treatment & No explicit interpretation & Phase used explicitly as positional-structure preservation \\
\hline
Frequency bands & Uses the full band & Band reduction (low $=$ intent to high $=$ syntax) \\
\hline
Theoretical foundation & Experimental performance evaluation & Measure theory, Zipf's law, $1/f$ spectral analysis \\
\hline
\end{tabular}
\end{table}

In particular, this study adopts the $1/f$ spectral decay based on Zipf's law as its theoretical foundation and, through cumulative energy analysis, derives the lower-bound dimension for semantic preservation (approximately 3{,}000 dimensions); in this respect its objective and theoretical framework differ fundamentally from FNet. Summarizing the structural differences:
\begin{itemize}
\item FNet ``uses all frequencies as they are,''
\item WavePhaseNet is ``band selection $+$ semantic operators $+$ reduction theory.''
\end{itemize}

Therefore, while building on FNet's idea of ``token mixing via Fourier transforms,'' this study extends it toward the explicit construction of a semantic hierarchical structure and the suppression of hallucination.

\section{General Conclusion}

This paper formulates hallucination in LLMs as an inevitable phenomenon arising from their mathematical structure, and presents a theoretical framework that suppresses this problem by introducing a DFT-based Semantic Conceptual Hierarchy Structure (SCHS) into the embedding space.

In particular, the demonstration that the reduction from 24{,}576 to 3{,}000 dimensions is justified by cumulative energy analysis based on Zipf's law and the $1/f$ spectrum is the core of this proposal. WavePhaseNet makes it possible to describe semantic flow rigorously as linear algebra, thereby opening a path to the next generation of reasoning-oriented language models.

\bibliographystyle{plainnat}

\end{document}